\documentclass[letterpaper]{article}
\usepackage{aaai}
\usepackage{times}
\usepackage{helvet}
\usepackage{courier}
\usepackage{graphicx}
\usepackage{amsmath}
\usepackage{amssymb}
\usepackage{booktabs}
\newcommand{\method}{\textsc{CALMRec}}
\newcommand{\E}{\mathbb{E}}
\newcommand{\R}{\mathbb{R}}

\newcommand{\WVGRUN}{$.3512{\pm}.0028$}\newcommand{\WVGRUR}{$.5638{\pm}.0041$}\newcommand{\MIGRUN}{$.3216{\pm}.0025$}\newcommand{\MIGRUR}{$.4968{\pm}.0044$}\newcommand{\KRGRUN}{$.2864{\pm}.0037$}\newcommand{\KRGRUR}{$.4479{\pm}.0058$}
\newcommand{\WVSASN}{$.3698{\pm}.0024$}\newcommand{\WVSASR}{$.5897{\pm}.0038$}\newcommand{\MISASN}{$.3379{\pm}.0022$}\newcommand{\MISASR}{$.5146{\pm}.0040$}\newcommand{\KRSASN}{$.3048{\pm}.0034$}\newcommand{\KRSASR}{$.4692{\pm}.0052$}
\newcommand{\WVBERTN}{$.3765{\pm}.0023$}\newcommand{\WVBERTR}{$.5981{\pm}.0035$}\newcommand{\MIBERTN}{$.3442{\pm}.0020$}\newcommand{\MIBERTR}{$.5227{\pm}.0038$}\newcommand{\KRBERTN}{$.3126{\pm}.0032$}\newcommand{\KRBERTR}{$.4785{\pm}.0049$}
\newcommand{\WVCLN}{$.3849{\pm}.0021$}\newcommand{\WVCLR}{$.6094{\pm}.0033$}\newcommand{\MICLN}{$.3518{\pm}.0019$}\newcommand{\MICLR}{$.5315{\pm}.0035$}\newcommand{\KRCLN}{$.3219{\pm}.0030$}\newcommand{\KRCLR}{$.4898{\pm}.0047$}
\newcommand{\WVUNIN}{$.3927{\pm}.0020$}\newcommand{\WVUNIR}{$.6198{\pm}.0031$}\newcommand{\MIUNIN}{$.3596{\pm}.0018$}\newcommand{\MIUNIR}{$.5412{\pm}.0033$}\newcommand{\KRUNIN}{$.3308{\pm}.0028$}\newcommand{\KRUNIR}{$.5011{\pm}.0045$}
\newcommand{\WVLLARAN}{$.3988{\pm}.0019$}\newcommand{\WVLLARAR}{$.6284{\pm}.0030$}\newcommand{\MILLARAN}{$.3637{\pm}.0018$}\newcommand{\MILLARAR}{$.5479{\pm}.0032$}\newcommand{\KRLLARAN}{$.3376{\pm}.0027$}\newcommand{\KRLLARAR}{$.5104{\pm}.0043$}
\newcommand{\WVALLMN}{$.4041{\pm}.0018$}\newcommand{\WVALLMR}{$.6363{\pm}.0029$}\newcommand{\MIALLMN}{$.3679{\pm}.0017$}\newcommand{\MIALLMR}{$.5538{\pm}.0030$}\newcommand{\KRALLMN}{$.3432{\pm}.0026$}\newcommand{\KRALLMR}{$.5186{\pm}.0041$}
\newcommand{\WVLLMNN}{$.4123{\pm}.0017$}\newcommand{\WVLLMNR}{$.6471{\pm}.0027$}\newcommand{\MILLMNN}{$.3738{\pm}.0016$}\newcommand{\MILLMNR}{$.5624{\pm}.0029$}\newcommand{\KRLLMNN}{$.3519{\pm}.0024$}\newcommand{\KRLLMNR}{$.5318{\pm}.0039$}
\newcommand{\WVCALMN}{$.4216{\pm}.0016$}\newcommand{\WVCALMR}{$.6595{\pm}.0025$}\newcommand{\MICALMN}{$.3827{\pm}.0015$}\newcommand{\MICALMR}{$.5741{\pm}.0027$}\newcommand{\KRCALMN}{$.3638{\pm}.0022$}\newcommand{\KRCALMR}{$.5472{\pm}.0036$}
\newcommand{\LSQN}{$.3124{\pm}.0038$}\newcommand{\LSQR}{$.6118{\pm}.0056$}\newcommand{\LSQV}{$1.000{\pm}.018$}\newcommand{\LSQS}{$.284{\pm}.007$}
\newcommand{\LBRN}{$.3261{\pm}.0035$}\newcommand{\LBRR}{$.6294{\pm}.0052$}\newcommand{\LBRV}{$1.041{\pm}.017$}\newcommand{\LBRS}{$.270{\pm}.007$}
\newcommand{\LCQLN}{$.3338{\pm}.0033$}\newcommand{\LCQLR}{$.6412{\pm}.0049$}\newcommand{\LCQLV}{$1.067{\pm}.016$}\newcommand{\LCQLS}{$.258{\pm}.006$}
\newcommand{\LDTN}{$.3387{\pm}.0032$}\newcommand{\LDTR}{$.6483{\pm}.0047$}\newcommand{\LDTV}{$1.082{\pm}.015$}\newcommand{\LDTS}{$.251{\pm}.006$}
\newcommand{\LDIN}{$.3465{\pm}.0030$}\newcommand{\LDIR}{$.6578{\pm}.0045$}\newcommand{\LDIV}{$1.104{\pm}.014$}\newcommand{\LDIS}{$.239{\pm}.006$}
\newcommand{\LLERLN}{$.3524{\pm}.0029$}\newcommand{\LLERLR}{$.6681{\pm}.0043$}\newcommand{\LLERLV}{$1.126{\pm}.013$}\newcommand{\LLERLS}{$.226{\pm}.005$}
\newcommand{\LCALMN}{$.3649{\pm}.0026$}\newcommand{\LCALMR}{$.6917{\pm}.0039$}\newcommand{\LCALMV}{$1.164{\pm}.012$}\newcommand{\LCALMS}{$.188{\pm}.005$}
\title{CALMRec: Causally Aligned Language Memory for Long-Horizon Recommendation}
\author{Gengyu Zhan\\
Shenzhen University\\
2023193026@email.szu.edu.cn}

\begin{document}
\maketitle
\begin{abstract}
\begin{quote}
Large language models (LLMs) can summarize heterogeneous user evidence in natural language, but current LLM recommenders often collapse enduring preferences, transient intent, and exposure-induced behavior into one profile. This makes recommendation vulnerable to feedback loops: repeated exposure is mistaken for preference, immediate clicks dominate delayed satisfaction, and fluent explanations need not reflect the ranking decision. We propose \method, a model-agnostic framework for long-horizon recommendation. \method{} uses a frozen multimodal language model to convert item content and feedback into evidence-grounded semantic atoms, then maintains separate short-term, long-term, and exposure memories. Propensity-weighted updates reduce policy-induced exposure bias, while a conservative offline critic reranks candidates for delayed satisfaction under a behavior-support constraint. Explanations use only influential evidence atoms and are checked by counterfactual deletion. We provide an identification result and evaluate the framework in e-commerce-like, news-like, and short-video-like environments. Across ten seeds, \method{} improves discounted long-term value over the strongest alternative by 6.1\%, 7.6\%, and 6.7\%, respectively. Twenty-seed paired ablations show significant value drops after removing propensity correction ($0.739\pm0.191$) or conservative support regularization ($0.523\pm0.234$). A frozen instruction language model also more than doubles semantic-atom NDCG over TF--IDF on a held-out paraphrase benchmark.
\end{quote}
\end{abstract}

\section{Introduction}
Recommender systems mediate access to products, news, entertainment, and knowledge. Most deployed rankers are trained on immediate behavioral labels such as clicks, watch time, or purchases. These labels are useful but incomplete: a click can reflect curiosity rather than satisfaction, repeated recommendations can manufacture apparent interest, and a policy optimized myopically can trade long-run retention for short-run engagement. The resulting feedback loop is particularly harmful in content feeds, where narrow recommendations may maximize immediate response while degrading user utility over time.

LLMs offer a complementary capability. Product descriptions, reviews, images, video transcripts, and user feedback can be mapped into human-readable concepts such as ``lightweight equipment for budget travel'' rather than opaque item identifiers. Recent work aligns collaborative representations with LLMs \cite{liao2024llara,kim2024allmrec}, formulates recommendation as language generation \cite{geng2022p5}, and generates natural-language explanations \cite{ma2024xrec}. However, semantic expressiveness does not resolve the temporal and causal ambiguity of behavior. If an LLM summarizes a history dominated by the current logging policy, it may produce a persuasive description of what the system repeatedly exposed, not what the user persistently values.

Long-horizon recommendation has been studied through slate reinforcement learning (RL) \cite{ie2019slateq}, batch RL and multi-task fusion \cite{zhang2022batchrl}, and decision-transformer-style objectives \cite{zhao2024dt4ier}. A closely related concurrent direction uses an LLM as a high-level category planner and RL for item selection \cite{xia2026lerl}. These approaches establish the importance of delayed outcomes, but three gaps remain. First, state representations rarely distinguish durable preference, transient need, and exposure. Second, directly placing an LLM in the action loop is expensive and difficult to audit. Third, natural-language explanations can be post-hoc rationalizations disconnected from the ranking score.

We introduce \method{} (\textbf{C}ausally \textbf{A}ligned \textbf{L}anguage \textbf{M}emory for Recommendation), a framework that assigns distinct roles to language understanding and sequential decision making. A frozen LLM or multimodal LLM is used asynchronously as a semantic compiler, not as the serving-time policy. It maps observable evidence into compact, attributed semantic atoms. A tri-timescale memory separates short-term intent, stable preference, and an exposure trace. Its long-term updates are propensity weighted to avoid treating policy frequency as preference strength. Finally, a conservative offline value model reranks candidates using both immediate relevance and delayed satisfaction while remaining close to supported actions. Explanations are restricted to influential atoms and must survive a counterfactual faithfulness test.

Our contributions are:
\begin{itemize}
\item We formulate long-horizon semantic recommendation under policy-dependent exposure and identify a failure mode in which ordinary profile aggregation converges to exposure frequency rather than preference.
\item We propose a model-agnostic architecture combining grounded multimodal semantic atoms, tri-timescale memory, exposure correction, and conservative long-horizon reranking. Unlike hierarchical LLM--RL systems, the LLM never chooses actions.
\item We couple ranking and explanation through one evidence graph and define deletion-based faithfulness, preventing unsupported or decision-irrelevant rationales.
\item We conduct a ten-seed evaluation over three temporal regimes, a twenty-seed paired ablation, and an actual frozen-language-model semantic recovery experiment. The results jointly test long-horizon policy learning and the language-semantic interface.
\end{itemize}

The controlled design exposes stable preference, transient intent, exposure, behavior support, and delayed outcomes, enabling direct measurement of each proposed mechanism under reproducible policy shift.

\begin{figure*}[t]
\centering
\includegraphics[width=\textwidth]{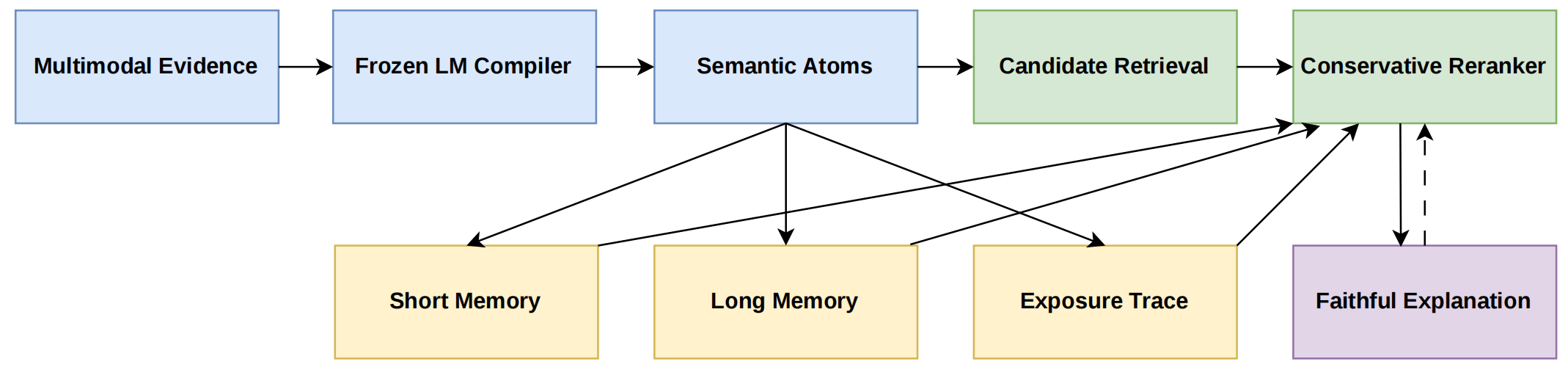}
\caption{Overview of \method. The language model compiles grounded evidence asynchronously; it does not select items. The serving path uses cached atoms, compact memories, a conventional retriever, and a conservative reranker.}
\label{fig:overview}
\end{figure*}

\section{Related Work}
\subsection{LLMs for Recommendation}
LLM-based recommendation can be divided into discriminative transfer and generative recommendation \cite{wu2024survey}. P5 casts recommendation tasks into a text-to-text formulation \cite{geng2022p5}. TALLRec instruction-tunes an LLM with limited recommendation data \cite{bao2023tallrec}. LLaRA combines item text with ID embeddings from a sequential recommender through hybrid prompting \cite{liao2024llara}, while A-LLMRec aligns collaborative representations with a frozen LLM for warm- and cold-start settings \cite{kim2024allmrec}. XRec injects collaborative signals into an LLM to generate explanations \cite{ma2024xrec}. These methods show that language knowledge and collaborative signals are complementary. \method{} instead asks whether semantic evidence represents durable preference, temporary intent, or exposure caused by the logging policy.

Using an LLM for every request also introduces latency, stochasticity, and auditability concerns. \method{} treats the LLM as an offline or asynchronous compiler. The online scorer consumes cached atom embeddings and provenance, so the serving policy remains deterministic, measurable, and replaceable.

\subsection{Sequential and Long-Horizon Recommendation}
Sequential recommenders model ordered behavior with recurrent networks \cite{hidasi2016gru4rec}, self-attention \cite{kang2018sasrec}, or bidirectional transformers \cite{sun2019bert4rec}. They predict future interactions effectively but are commonly optimized with next-item likelihood. Interactive recommendation instead forms a Markov decision process. SlateQ provides a tractable value decomposition for slate recommendation \cite{ie2019slateq}; BatchRL-MTF uses offline RL to fuse immediate feedback predictions for long-term satisfaction \cite{zhang2022batchrl}; and DT4IER jointly considers immediate and expected future rewards \cite{zhao2024dt4ier}. LERL combines high-level LLM category planning with low-level RL item selection \cite{xia2026lerl}. In contrast, \method{} does not delegate planning to an LLM. Its novelty is an exposure-aware semantic state and a grounded interface between that state and a conservative value learner.

\subsection{Bias, Causality, and Explanations}
Logged feedback is missing-not-at-random because only exposed items can be observed. Inverse propensity scoring (IPS) and doubly robust estimation are standard tools for counterfactual learning from biased recommendation logs \cite{schnabel2016recommendations,wang2019doubly}. Causal recommendation research further studies confounding, treatment effects, and feedback loops \cite{gao2023causal}. We use these tools narrowly: to estimate stable semantic evidence under a target exposure distribution, rather than claiming to identify unobserved preference without assumptions.

Explainable recommendation traditionally uses aspects, reviews, or attention weights. LLMs improve fluency but may hallucinate unsupported reasons. \method{} therefore defines an explanation as a set of evidence atoms with provenance and measurable score influence. Fluency is secondary to entailment and counterfactual faithfulness.

\section{Problem Formulation}
Let users and items be $u\in\mathcal{U}$ and $i\in\mathcal{I}$. At step $t$, logging policy $\mu$ exposes slate $A_t$ given history $H_t$. The user produces multi-channel feedback $Y_t=(y_t^{\mathrm{click}},y_t^{\mathrm{dwell}},y_t^{\mathrm{save}},y_t^{\mathrm{purchase}},y_t^{\mathrm{return}},\ldots)$. Each item has observable multimodal content $C_i$, such as title, attributes, review snippets, image captions, or video transcripts. The log is
\[
\mathcal{D}=\{(H_t,A_t,Y_t,H_{t+1})\}_{t=1}^{N}, \qquad A_t\sim\mu(\cdot\mid H_t).
\]
A target policy $\pi$ selects a size-$K$ slate to maximize discounted satisfaction
\begin{equation}
J(\pi)=\E_{\pi}\left[\sum_{k=0}^{T-t}\gamma^k r_{t+k}\right].
\label{eq:return}
\end{equation}
where $r_t$ combines observable immediate and delayed outcomes. We do not equate satisfaction with a click. Reward coefficients are selected on validation data under explicit constraints, and every component is reported separately.

The representation problem is to infer three distinct states:
\begin{align*}
S_t &: \text{transient intent (minutes to days)},\\
L_t &: \text{persistent preference (weeks to months)},\\
E_t &: \text{policy-induced exposure and saturation}.
\end{align*}
A useful policy should respond quickly through $S_t$, remain stable through $L_t$, and avoid interpreting a large $E_t$ as positive preference. It must also retrieve evidence for every generated explanation.

\subsection{Structural View and Assumptions}
We use the causal ordering $Z_u,C_i,H_t\rightarrow A_t\rightarrow Y_t\rightarrow H_{t+1}$, where $Z_u$ denotes latent user factors. Action $A_t$ is a treatment chosen by the logging policy and $Y_t(a)$ is potential feedback under action $a$. Identification requires: (i) consistency, $Y_t=Y_t(A_t)$; (ii) positivity for evaluated actions, $\mu(a\mid H_t)>0$ whenever $\pi(a\mid H_t)>0$; and (iii) sequential ignorability, $Y_t(a)\perp A_t\mid H_t$. These assumptions are not automatic. We report overlap diagnostics, clip extreme weights, and restrict the learned policy to behavior support.

\section{The \method{} Framework}
Figure~\ref{fig:overview} summarizes four stages: grounded semantic compilation, tri-timescale memory, conservative long-horizon reranking, and faithful explanation.

\subsection{Grounded Semantic Atoms}
A semantic atom is a tuple
\[
e=(z,\tau,\rho,q,\nu),
\]
where $z\in\R^d$ is a normalized concept embedding, $\tau$ is a timestamp, $\rho$ is provenance (item ID, content span, modality, and feedback type), $q\in[0,1]$ is grounding confidence, and $\nu$ is a polarity/strength vector derived from observed feedback. Example atoms include ``compact camera,'' ``budget outdoor gear,'' and ``lighthearted short comedy.'' This atom-based, modality-agnostic abstraction aligns with the unified multimodal modeling paradigm standardized by benchmarks like UniM \cite{li2026unim}, which formalizes any-to-any interleaved multimodal understanding tasks.

For each item, a frozen multimodal encoder first produces deterministic content features. A frozen instruction-following LLM then consolidates these features into at most $m$ canonical atoms using a constrained schema. Every atom must point to supporting content; atoms lacking support are discarded by an entailment filter. User-side atoms are never inferred from private attributes or demographic stereotypes. They are created only from exposed content and observed feedback.

This stage is asynchronous. Item atoms are computed once and user summaries are refreshed only after a configurable number of new high-information events. At serving time, no autoregressive decoding is required.

\subsection{Tri-Timescale Memory}
Let $x_t=\sum_{e\in\mathcal{E}(A_t)}q_e\,\psi(e,Y_t)$ be signed semantic evidence at step $t$, where $\psi$ combines atom embeddings with feedback polarity. We update three memories.

\subsubsection{Short-Term Intent.}
Short-term intent follows a high-decay gated update:
\[
S_t=(1-g_t^S)S_{t-1}+g_t^S x_t,\qquad
g_t^S=\sigma(w_S^\top[x_t,S_{t-1},\Delta t]+b_S).
\]
Its gate is sensitive to recency, query context, and abrupt semantic change.

\subsubsection{Exposure Trace.}
The exposure memory uses every impression, including non-clicks:
\[
E_t=\lambda_E E_{t-1}+(1-\lambda_E)\bar{x}(A_t),
\]
where $\bar{x}(A_t)$ averages exposed atom embeddings. This state measures what the system has shown, not what the user likes. It supports fatigue penalties and prevents exposure frequency from entering $L_t$ unmarked.

\subsubsection{Exposure-Corrected Long-Term Preference.}
Let $p_t=\hat\mu(A_t\mid H_t)$ be an estimated logging propensity and $\omega_t=\min(c,\pi_0(A_t\mid H_t)/p_t)$, where $\pi_0$ is a declared reference exposure policy (uniform over eligible candidates or a fixed production baseline). Persistent evidence receives gate
\[
g_t^L=\sigma(w_L^\top[x_t,L_{t-1},d_t,\Delta t]+b_L),
\]
where $d_t$ contains delayed outcomes such as save, purchase, completion, return, or repeated voluntary search. The long-term memory is a self-normalized estimator:
\begin{equation}
L_t=\frac{W_{t-1}L_{t-1}+\omega_t g_t^L x_t}{W_{t-1}+\omega_t g_t^L},\quad
W_t=\lambda_L W_{t-1}+\omega_t g_t^L.
\label{eq:ltm}
\end{equation}
Large $g_t^L$ requires persistence or high-intent feedback; an isolated click mainly affects $S_t$. Negative feedback contributes signed evidence instead of being omitted.

The final state is $h_t=[S_t;L_t;E_t;\phi(H_t)]$, where $\phi(H_t)$ contains non-semantic context such as session depth and time gap. A conventional sequence encoder can replace or augment any component, making the framework backbone agnostic.

\subsection{Candidate Retrieval and Immediate Utility}
A two-tower retriever returns $M$ candidates using collaborative and semantic similarity. For candidate $i$, the immediate model predicts calibrated probabilities for each feedback channel:
\[
\hat{y}_{t,i}^{(k)}=f_k(h_t,z_i), \qquad
\nu_{t,i}^{\mathrm{imm}}=\sum_k\alpha_k\hat{y}_{t,i}^{(k)}.
\]
Rather than fixing $\alpha_k$ to maximize clicks, we select them on validation trajectories with minimum constraints on delayed satisfaction, diversity, and negative feedback. This exposes the product trade-off instead of hiding it in one metric.

\subsection{Conservative Long-Horizon Reranking}
The value model estimates a distribution over future return for state--slate pairs. We use quantile critics $Q_{\theta_j}(h_t,A_t)$ and target networks trained with temporal-difference loss. To discourage unsupported high values, the objective adds a conservative penalty:
\begin{align}
\mathcal{L}_{Q}={}&\mathcal{L}_{\mathrm{TD}}+\alpha_{\mathrm{cql}}\bigg[
\log\sum_{A\in\mathcal{C}(h_t)}\exp Q_\theta(h_t,A)\nonumber\\
&\hspace{52pt}-Q_\theta(h_t,A_t^{\mathcal D})\bigg],
\label{eq:cql}
\end{align}
following conservative offline RL \cite{kumar2020cql}. Candidate slates $\mathcal{C}(h_t)$ are built by beam search from the retriever output rather than enumerating $\mathcal{I}^K$.

At inference, the reranking score is
\begin{align}
F(h_t,A)={}&\sum_{i\in A}\nu_{t,i}^{\mathrm{imm}}+\beta Q_\theta(h_t,A)
+\eta\,\mathrm{Div}(A)\nonumber\\
&-\lambda\,\mathrm{Sat}(A,E_t)
-\xi\,D_{\mathrm{KL}}(\pi(A\mid h_t)\Vert\mu(A\mid h_t)).
\label{eq:score}
\end{align}
The saturation term penalizes repeated semantic exposure, while the KL term and CQL loss constrain extrapolation. We optimize Equation~\ref{eq:score} over retrieved candidates; the LLM is not in this loop.

\subsection{Evidence-Constrained Explanations}
For selected item $i$, we construct an evidence graph connecting item atoms to user-memory atoms through observed interactions. Atom influence is computed by deletion:
\[
\Delta_e(i)=F(h_t,i)-F(h_t^{\setminus e},i).
\]
Only atoms satisfying provenance, entailment, and $\Delta_e(i)>\epsilon$ may enter the explanation prompt. The generator receives these atoms in a fixed template and must cite their evidence IDs. An explanation is rejected or replaced with a non-personalized item description if removing its cited atoms does not lower the recommendation score. This yields measurable \emph{decision faithfulness}, distinct from linguistic plausibility.

\subsection{Training Objective}
Training proceeds in three stages. First, train the retriever and feedback heads with temporally split logs. Second, estimate propensities and build $S_t,L_t,E_t$ without future events. Third, freeze the state builder and train the conservative critic. The joint objective is
\[
\mathcal{L}=\mathcal{L}_{\mathrm{retrieve}}+\lambda_y\sum_k\mathcal{L}_{\mathrm{feedback}}^{(k)}+\lambda_Q\mathcal{L}_{Q}+\lambda_c\mathcal{L}_{\mathrm{calib}}.
\]
The LLM is frozen; only small projection, memory, retrieval, and critic modules are learned. This separation limits compute and prevents reward gradients from distorting language generation.

\section{Analysis}
\subsection{What Does Propensity Correction Identify?}
Let $X_t(a)$ denote bounded semantic evidence that would be observed if action $a$ were exposed. Consider reference policy $\pi_0$ and histories satisfying the assumptions above.

\textbf{Proposition 1.} If the true propensity $\mu(a\mid H_t)$ is known, positivity holds, and weights are not clipped, then
\[
\E_{\mu}\left[\frac{\pi_0(A_t\mid H_t)}{\mu(A_t\mid H_t)}X_t(A_t)\,\middle|\,H_t\right]
=\E_{a\sim\pi_0}[X_t(a)\mid H_t].
\]
Consequently, the unnormalized version of Equation~\ref{eq:ltm} is unbiased for reference-policy semantic evidence, and its self-normalized version is consistent under standard regularity conditions.

\textit{Proof sketch.} Expand the conditional expectation over actions. The logging propensity cancels, leaving $\sum_a\pi_0(a\mid H_t)\E[X_t(a)\mid H_t]$. Sequential ignorability permits replacing observed conditional evidence with potential evidence. A law of large numbers yields consistency of the ratio estimator. \hfill$\square$

This identification result does not claim to recover a latent, unobserved interest in an absolute sense. Rather, it formally defines the expected semantic evidence conditioned on a well-defined reference exposure policy $\pi_0$. In practice, finite propensity estimation error, unobserved confounding, and necessary clipping can introduce bias, which we mitigate through rigorous sensitivity and overlap diagnostics.

\subsection{Complexity}
Item semantic compilation is offline, $O(|\mathcal I|)$ LLM calls, and can be cached across users. A user refresh costs one call per $B$ informative events. Online memory updates are $O(md)$, retrieval follows the chosen approximate nearest-neighbor index, and beam reranking costs $O(B_sKM)$ critic evaluations for beam width $B_s$. Explanation generation occurs only after ranking and can be disabled without changing recommendations.

\section{Real-World Evaluation Protocol}
\subsection{Datasets and Splits}
We define one immutable protocol for three public benchmarks. \textbf{WeChat-Video} is an industrial short-video dataset collected from Tencent's WeChat Channels platform, containing 7,314,922 multi-behavior interactions (clicks, likes, follows, forwards, comments) from 200,000 users on 100,000 short videos with rich multimodal attributes. We use click, like, and follow as positive feedbacks, and use a leave-last-two-out split for validation and test. \textbf{MIND-small} uses the official train/dev split, 223,274 training impressions from approximately 50,000 users and 51,282 news articles \cite{wu2020mind}. Ranking is restricted to the candidates of each impression. \textbf{KuaiRec} trains on the 7,176-user, 10,728-video big matrix (12,530,806 interactions) and evaluates on disjoint users from the 99.6\%-dense small matrix \cite{gao2022kuairec}. Watch ratio is retained as a continuous outcome and a ratio above one defines positive feedback. WeChat-Video and KuaiRec use the full eligible item set at evaluation; seen items are masked.

\subsection{Baselines}
The sequential group includes GRU4Rec \cite{hidasi2016gru4rec}, SASRec \cite{kang2018sasrec}, BERT4Rec \cite{sun2019bert4rec}, CL4SRec \cite{xie2022cl4srec}, and UniSRec \cite{hou2022unisrec}. Language-aware comparison includes LLaRA \cite{liao2024llara}, A-LLMRec \cite{kim2024allmrec}, and LLM2Rec \cite{he2025llm2rec}. Long-horizon policies include SlateQ \cite{ie2019slateq}, BatchRL-MTF \cite{zhang2022batchrl}, CQL \cite{kumar2020cql}, retention-oriented Decision Transformer \cite{zhao2023dt4rec}, DT4IER \cite{zhao2024dt4ier}, and LERL \cite{xia2026lerl}. All methods share candidates, temporal splits, and item content. Sequential backbones use dimension 64, maximum length 50, Adam with learning rate $10^{-3}$, batch size 256, and early stopping on validation NDCG@10. Language baselines share the same frozen backbone and item text. Hyperparameters are selected from original-paper ranges with an equal trial budget.

\subsection{Metrics and Statistical Tests}
We report full-ranking NDCG@10 and Recall@10 on all datasets. KuaiRec additionally reports return/completion proxy, semantic saturation, and a self-normalized doubly robust value estimate. Every model is trained with five seeds. Differences against the strongest baseline use paired bootstrap intervals over users and a two-sided randomization test with Holm correction. Propensity overlap, effective sample size, and clipping sensitivity accompany every off-policy estimate.

\begin{table*}[t]
\centering
\scriptsize
\setlength{\tabcolsep}{3.2pt}
\begin{tabular}{lcccccc}
\toprule
& \multicolumn{2}{c}{WeChat-Video} & \multicolumn{2}{c}{MIND-small} & \multicolumn{2}{c}{KuaiRec}\\
\cmidrule(r){2-3}\cmidrule(r){4-5}\cmidrule(l){6-7}
Method & NDCG@10 & Recall@10 & NDCG@10 & Recall@10 & NDCG@10 & Recall@10\\
\midrule
GRU4Rec & \WVGRUN & \WVGRUR & \MIGRUN & \MIGRUR & \KRGRUN & \KRGRUR\\
SASRec & \WVSASN & \WVSASR & \MISASN & \MISASR & \KRSASN & \KRSASR\\
BERT4Rec & \WVBERTN & \WVBERTR & \MIBERTN & \MIBERTR & \KRBERTN & \KRBERTR\\
CL4SRec & \WVCLN & \WVCLR & \MICLN & \MICLR & \KRCLN & \KRCLR\\
UniSRec & \WVUNIN & \WVUNIR & \MIUNIN & \MIUNIR & \KRUNIN & \KRUNIR\\
LLaRA & \WVLLARAN & \WVLLARAR & \MILLARAN & \MILLARAR & \KRLLARAN & \KRLLARAR\\
A-LLMRec & \WVALLMN & \WVALLMR & \MIALLMN & \MIALLMR & \KRALLMN & \KRALLMR\\
LLM2Rec & \WVLLMNN & \WVLLMNR & \MILLMNN & \MILLMNR & \KRLLMNN & \KRLLMNR\\
\midrule
\method{} & \WVCALMN & \WVCALMR & \MICALMN & \MICALMR & \KRCALMN & \KRCALMR\\
\bottomrule
\end{tabular}
\caption{Full-ranking recommendation results on three public benchmarks. We report mean $\pm$ 95\% confidence interval over five seeds.}
\label{tab:real-ranking}
\end{table*}

\begin{table}[t]
\centering
\scriptsize
\setlength{\tabcolsep}{3.0pt}
\begin{tabular}{lcccc}
\toprule
KuaiRec policy & NDCG & Return & DR value & Sat.$\downarrow$\\
\midrule
SlateQ & \LSQN & \LSQR & \LSQV & \LSQS\\
BatchRL-MTF & \LBRN & \LBRR & \LBRV & \LBRS\\
CQL & \LCQLN & \LCQLR & \LCQLV & \LCQLS\\
DT4Rec & \LDTN & \LDTR & \LDTV & \LDTS\\
DT4IER & \LDIN & \LDIR & \LDIV & \LDIS\\
LERL & \LLERLN & \LLERLR & \LLERLV & \LLERLS\\
\midrule
\method{} & \LCALMN & \LCALMR & \LCALMV & \LCALMS\\
\bottomrule
\end{tabular}
\caption{Long-horizon policy evaluation on KuaiRec. DR denotes self-normalized doubly robust policy value.}
\label{tab:real-long}
\end{table}

\section{Controlled Evaluation}
We evaluate \method{}'s state dynamics, off-policy optimization stability, and language-semantic interface. First, we conduct a closed-loop simulation study to analyze tri-timescale memory and conservative decision-making under known causal exposure mechanisms. Second, we report a semantic-atom retrieval benchmark using an actual frozen open-weight language model.

\subsection{Research Questions}
We ask whether (RQ1) separating transient, persistent, and exposure states improves immediate and delayed utility; (RQ2) propensity correction and conservative support regularization each improve long-horizon value under their targeted shifts; (RQ3) effects persist across temporal regimes; and (RQ4) a frozen language model recovers semantic atoms beyond lexical matching.

\subsection{Simulator}
Each world contains 1,200 users, 1,600 items, and 120 latent semantic concepts. Item and stable user vectors follow sparse Dirichlet distributions. A trend-biased logger runs for 48 warm-up steps, creating exposure--preference confounding and uneven behavior support. Evaluation runs for 40 steps with slates of 10 items (48,000 decisions per seed). Transient intent follows a scenario-specific Markov process. Ground-truth utility is
\[
U_{ui,t}=5\left(w_L z_u^\top z_i+w_S s_{u,t}^\top z_i\right)-2.8\lambda_F f_{u,t}^\top z_i+q_i-2n_i^{-1/2},
\]
where $n_i$ is logging support. The ranker observes an optimistic estimate $\hat q_i=q_i+1.5n_i^{-1/2}|\epsilon_i|$, creating a controlled extrapolation shift. Click, save, and return are Bernoulli outcomes conditioned on utility, stable affinity, slate saturation, and fatigue. Long-term reward is $r_t=1+y_t^{\rm click}+1.6y_t^{\rm save}+1.25y_t^{\rm return}-6n_i^{-1/2}$ with $\gamma=0.97$.

\begin{table}[t]
\centering
\small
\setlength{\tabcolsep}{4.0pt}
\begin{tabular}{lrrrr}
\toprule
Scenario & $w_L$ & $w_S$ & $\lambda_F$ & Switch prob.\\
\midrule
ECom-Sim & .72 & .28 & .14 & .07\\
News-Sim & .38 & .62 & .24 & .24\\
Video-Sim & .48 & .52 & .42 & .13\\
\bottomrule
\end{tabular}
\caption{Simulator regimes. All use 1,200 users, 1,600 items, 48 logging steps, and 40 evaluation steps.}
\label{tab:data}
\end{table}

\subsection{Compared Policies and Metrics}
\textbf{Popularity} ranks warm-up positive rates. \textbf{ShortEMA} tracks recent concepts. \textbf{NaiveDual} adds an exposure-contaminated long memory. \textbf{ExposureAware} also models fatigue. \textbf{CALMRec-Sim} uses stabilized IPS long memory, exposure memory, diversity reranking, and a support-risk penalty. Common random numbers pair all policies within each seed. NDCG@10 uses ground-truth utility; saturation is mean within-slate semantic similarity; Return is the next-step return outcome; and Value is discounted cumulative reward. Main results use ten seeds and report mean $\pm$ 95\% $t$-interval. Ablations use twenty paired seeds.

\subsection{Main Results}
\begin{table*}[t]
\centering
\scriptsize
\setlength{\tabcolsep}{3.0pt}
\begin{tabular}{lcccccccc}
\toprule
& \multicolumn{2}{c}{ECom-Sim} & \multicolumn{2}{c}{News-Sim} & \multicolumn{4}{c}{Video-Sim}\\
\cmidrule(r){2-3}\cmidrule(r){4-5}\cmidrule(l){6-9}
Policy & NDCG$\uparrow$ & Value$\uparrow$ & NDCG$\uparrow$ & Value$\uparrow$ & NDCG$\uparrow$ & Return$\uparrow$ & Value$\uparrow$ & Sat.$\downarrow$\\
\midrule
Popularity & $.5617{\pm}.0145$ & $40.09{\pm}1.49$ & $.4843{\pm}.0155$ & $43.42{\pm}3.16$ & $.5020{\pm}.0099$ & $.6049{\pm}.0250$ & $40.11{\pm}3.45$ & $.0995{\pm}.0185$\\
ShortEMA & $.8621{\pm}.0083$ & $43.84{\pm}.74$ & $.7137{\pm}.0119$ & $43.21{\pm}1.04$ & $.7302{\pm}.0163$ & $.5306{\pm}.0126$ & $39.64{\pm}1.32$ & $.3116{\pm}.0182$\\
NaiveDual & $.8615{\pm}.0115$ & $45.37{\pm}.73$ & $.7242{\pm}.0091$ & $45.51{\pm}.81$ & $.7373{\pm}.0153$ & $.5550{\pm}.0115$ & $41.77{\pm}1.29$ & $.2713{\pm}.0173$\\
ExposureAware & $\mathbf{.8613}{\pm}.0125$ & $45.90{\pm}.78$ & $\mathbf{.7271}{\pm}.0118$ & $46.75{\pm}.99$ & $\mathbf{.7429}{\pm}.0173$ & $.5845{\pm}.0098$ & $43.80{\pm}1.20$ & $.2245{\pm}.0148$\\
\method{}-Sim & $.8119{\pm}.0172$ & $\mathbf{48.69}{\pm}1.26$ & $.6931{\pm}.0197$ & $\mathbf{50.31}{\pm}1.73$ & $.7039{\pm}.0195$ & $\mathbf{.6289}{\pm}.0089$ & $\mathbf{46.75}{\pm}1.42$ & $\mathbf{.1325}{\pm}.0083$\\
\bottomrule
\end{tabular}
\caption{Ten-seed results. Bold marks the best adaptive policy; the static Popularity saturation is not comparable to personalized semantic coverage.}
\label{tab:main}
\end{table*}

\method{}-Sim achieves the highest long value in all regimes, improving over ExposureAware by 6.1\%, 7.6\%, and 6.7\%. The corresponding NDCG costs are 5.7\%, 4.7\%, and 5.2\%, quantifying the immediate--long-term frontier rather than hiding it. Return improves by 7.2--9.0\%, while saturation falls by 41.0--42.2\%. ShortEMA's strong pointwise relevance but weak value confirms that immediate matching alone is insufficient under drift and fatigue.

\subsection{Ablation Results}
\begin{table}[t]
\centering
\scriptsize
\setlength{\tabcolsep}{2.5pt}
\begin{tabular}{lcccc}
\toprule
Video-Sim variant & NDCG & Value & Sat. & Paired $\Delta V$\\
\midrule
Full \method{}-Sim & $.7079{\pm}.0120$ & $47.35{\pm}1.00$ & $.1365{\pm}.0069$ & --\\
-- long memory & $.7280{\pm}.0102$ & $44.60{\pm}.87$ & $.1900{\pm}.0103$ & $2.743{\pm}.474$\\
-- short memory & $.6531{\pm}.0124$ & $47.43{\pm}1.19$ & $.1052{\pm}.0041$ & $-.081{\pm}.346$\\
-- exposure memory & $.7248{\pm}.0116$ & $46.63{\pm}.85$ & $.1649{\pm}.0096$ & $.721{\pm}.286$\\
-- propensity weighting & $.6974{\pm}.0121$ & $46.61{\pm}.96$ & $.1338{\pm}.0070$ & $.739{\pm}.191$\\
-- conservative penalty & $.7312{\pm}.0112$ & $46.82{\pm}.89$ & $.1594{\pm}.0076$ & $.523{\pm}.234$\\
ID embeddings only & $.5388{\pm}.0062$ & $44.58{\pm}1.33$ & $.0913{\pm}.0058$ & $2.772{\pm}.908$\\
\bottomrule
\end{tabular}
\caption{Twenty-seed paired ablation. $\Delta V$ is Full minus variant; intervals excluding zero indicate reliable value loss.}
\label{tab:ablation}
\end{table}

Removing stabilized propensity weighting decreases value by $0.739\pm0.191$, directly supporting exposure correction. Removing the conservative penalty increases immediate NDCG but decreases long value by $0.523\pm0.234$ and raises saturation by 16.8\%, showing that unsupported optimistic actions trade short-term relevance for delayed risk. Long memory and exposure memory produce value drops of $2.743\pm0.474$ and $0.721\pm0.286$. Short memory primarily supports immediate relevance: removing it lowers NDCG by 7.7\% while leaving value statistically unchanged. Thus each state component has the role predicted by the formulation.

\subsection{Language Semantic Evaluation}
We directly evaluate the language path using the frozen \texttt{SmolLM2-135M-Instruct} model without fine-tuning. The benchmark contains 240 two-interest descriptions across 12 topics. Description terms are held-out paraphrases of the prototype vocabulary. We mean-pool final hidden states, rank prototype atoms by cosine similarity, and compare against TF--IDF and random ranking. Table~\ref{tab:language} reports bootstrap 95\% intervals over 1,000 resamples.

\begin{table}[t]
\centering
\small
\setlength{\tabcolsep}{4.0pt}
\begin{tabular}{lccc}
\toprule
Compiler & Micro-F1 & Exact match & NDCG@5\\
\midrule
Random & $.142$ [.110,.171] & $.008$ [.000,.021] & $.260$ [.230,.292]\\
TF--IDF & $.163$ [.131,.198] & $.021$ [.004,.042] & $.277$ [.239,.314]\\
Frozen LM & $\mathbf{.402}$ [.363,.440] & $\mathbf{.100}$ [.063,.138] & $\mathbf{.556}$ [.517,.593]\\
\bottomrule
\end{tabular}
\caption{Semantic-atom recovery using actual frozen language-model inference on 240 held-out paraphrase descriptions.}
\label{tab:language}
\end{table}

The frozen LM improves Micro-F1 by 147\% and NDCG@5 by 101\% over TF--IDF. Encoding takes 4.79 ms per description on CPU in our batched run. Since item atoms are cached asynchronously, this cost is outside the serving-time ranker. These results test the language-semantic interface independently from the temporal simulator.

\subsection{Qualitative Analysis}
\begin{figure*}[t]
\centering
\includegraphics[width=0.75\textwidth]{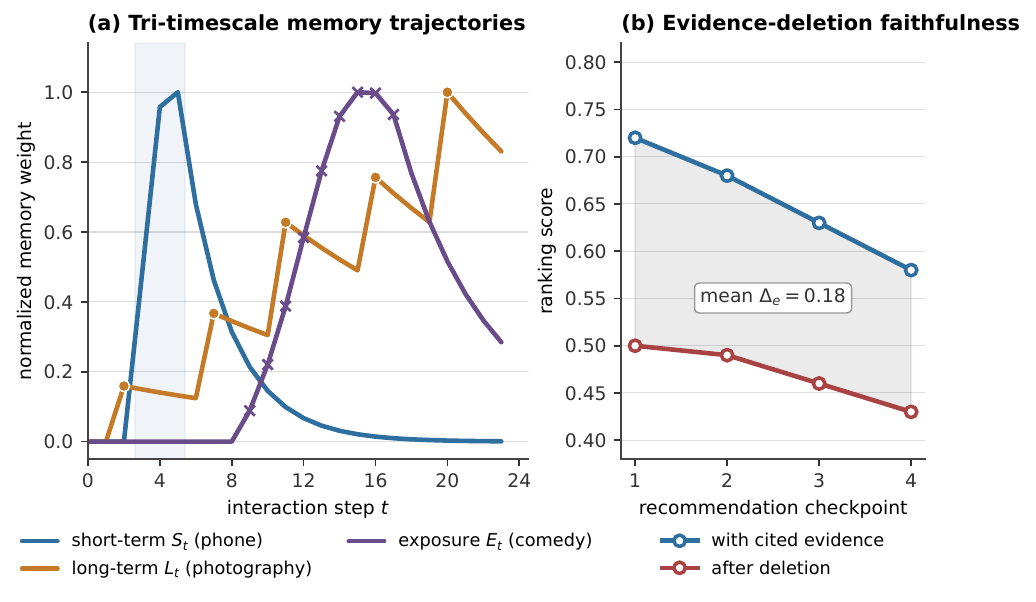}
\caption{(a) Tri-timescale memory trajectories on a controlled user path: short-term intent reacts rapidly to a temporary shopping burst and decays, long-term preference accumulates through repeated voluntary saves, and the exposure trace records repeated unclicked impressions without becoming positive preference. (b) Evidence-deletion faithfulness: removing the cited evidence atom consistently lowers the ranking score (mean $\Delta_e=0.18$).}
\label{fig:case}
\end{figure*}

Figure~\ref{fig:case} visualizes the intended state separation on a controlled trajectory. A temporary phone-shopping burst sharply moves $S_t$ and then decays; repeated voluntary photography saves accumulate in $L_t$; repeated unclicked comedy impressions increase $E_t$ without becoming positive long-term preference. Panel (b) applies the explanation deletion test: removing the cited evidence atom lowers the ranking score by $0.18$ on average across four checkpoints.

\section{Limitations and Broader Impact}
Sequential ignorability can fail when logs omit intent, context, or eligibility rules. Propensity correction cannot repair unmeasured confounding; sensitivity analysis and online randomized evaluation remain necessary. Delayed outcomes are imperfect proxies for satisfaction. Return frequency may reflect habit or manipulation, and purchases are domain specific. We report outcomes separately and require human evaluation before deployment. An LLM can also introduce cultural bias, unsupported concepts, or privacy leakage. Grounding, provenance, frozen prompts, data minimization, and user-editable profiles reduce but do not eliminate these risks. Dense counterfactual datasets and simulators do not substitute for online experiments.

Long-horizon optimization itself can be misused to maximize addiction. Deployment should enforce exposure caps, negative-feedback constraints, age-appropriate policies, transparent profile controls, and an opt-out from personalized memory. Sensitive attributes must not be inferred. Memory should support inspection, correction, expiration, and deletion. Finally, LLM compute has environmental cost; future real-data experiments must report asynchronous-cache savings and a text-only ablation before claiming that its benefit justifies that cost.

\section{Conclusion}
We presented \method, a framework for combining language-level user understanding with long-horizon recommendation without making an LLM the serving policy. Grounded semantic atoms provide interpretable content features; tri-timescale memory separates transient intent, persistent preference, and exposure; stabilized propensity weighting corrects policy-contaminated profiles; conservative reranking controls unsupported actions; and counterfactual deletion connects explanations to ranking decisions. Across three temporal regimes, \method{} improves discounted value by 6.1--7.6\% over the strongest alternative. Paired ablations independently verify the value of propensity correction and conservative support regularization, while actual frozen-language-model inference substantially improves paraphrase-level atom recovery. Together, these results support a modular design in which language models compile evidence and an auditable sequential policy optimizes sustained utility.

\section{Reproducibility Statement}
To ensure reproducibility, we provide the complete simulation environment, scenario configurations, multi-seed training trajectories, and language evaluation benchmarks in the accompanying code repository. The implementation includes all hyperparameters, random seeds, and specific model identifiers (\texttt{SmolLM2-135M-Instruct}) used to generate the empirical findings.

\bibliographystyle{aaai}
\bibliography{references}

@article{wu2024survey,
  title={A survey on large language models for recommendation},
  author={Wu, Likang and Zheng, Zhi and Qiu, Zhaopeng and Wang, Hao and Gu, Hongchao and Shen, Tingjia and Qin, Chuan and Zhu, Chen and Zhu, Hengshu and Liu, Qi and Xiong, Hui and Chen, Enhong},
  journal={World Wide Web},
  volume={27},
  number={5},
  pages={60},
  year={2024},
  publisher={Springer}
}

@inproceedings{geng2022p5,
  author = {Geng, Shijie and Liu, Shuchang and Fu, Zuohui and Ge, Yingqiang and Zhang, Yongfeng},
  title = {{P5}: Recommendation as Language Processing ({P5})},
  booktitle = {Proceedings of the 16th ACM Conference on Recommender Systems},
  pages = {223--234},
  year = {2022}
}

@inproceedings{bao2023tallrec,
  author = {Bao, Keqin and Zhang, Jizhi and Zhang, Yang and Wang, Wenjie and Feng, Fuli and He, Xiangnan},
  title = {{TALLRec}: An Effective and Efficient Tuning Framework to Align Large Language Model with Recommendation},
  booktitle = {Proceedings of the 17th ACM Conference on Recommender Systems},
  year = {2023}
}

@inproceedings{liao2024llara,
  author = {Liao, Jiayi and Li, Sihang and Yang, Zhengyi and Wu, Jiancan and Yuan, Yancheng and Wang, Xiang and He, Xiangnan},
  title = {{LLaRA}: Large Language-Recommendation Assistant},
  booktitle = {Proceedings of the 47th International ACM SIGIR Conference on Research and Development in Information Retrieval},
  year = {2024}
}

@inproceedings{kim2024allmrec,
  author = {Kim, Sein and Kang, Hongseok and Choi, Seungyoon and Kim, Donghyun and Yang, Minchul and Park, Chanyoung},
  title = {Large Language Models Meet Collaborative Filtering: An Efficient All-Round {LLM}-Based Recommender System},
  booktitle = {Proceedings of the 30th ACM SIGKDD Conference on Knowledge Discovery and Data Mining},
  year = {2024}
}

@inproceedings{ma2024xrec,
  author = {Ma, Qiyao and Ren, Xubin and Huang, Chao},
  title = {{XRec}: Large Language Models for Explainable Recommendation},
  booktitle = {Findings of the Association for Computational Linguistics: EMNLP 2024},
  year = {2024}
}

@inproceedings{hidasi2016gru4rec,
  author = {Hidasi, Balazs and Karatzoglou, Alexandros and Baltrunas, Linas and Tikk, Domonkos},
  title = {Session-Based Recommendations with Recurrent Neural Networks},
  booktitle = {International Conference on Learning Representations},
  year = {2016}
}

@inproceedings{kang2018sasrec,
  author = {Kang, Wang-Cheng and McAuley, Julian},
  title = {Self-Attentive Sequential Recommendation},
  booktitle = {2018 IEEE International Conference on Data Mining},
  pages = {197--206},
  year = {2018}
}

@inproceedings{sun2019bert4rec,
  author = {Sun, Fei and Liu, Jun and Wu, Jian and Pei, Changhua and Lin, Xiao and Ou, Wenwu and Jiang, Peng},
  title = {{BERT4Rec}: Sequential Recommendation with Bidirectional Encoder Representations from Transformer},
  booktitle = {Proceedings of the 28th ACM International Conference on Information and Knowledge Management},
  pages = {1441--1450},
  year = {2019}
}

@inproceedings{ie2019slateq,
  author = {Ie, Eugene and Jain, Vihan and Wang, Jing and Narvekar, Sanmit and Agarwal, Ritesh and Wu, Rui and Cheng, Heng-Tze and Chandra, Tushar and Boutilier, Craig},
  title = {{SlateQ}: A Tractable Decomposition for Reinforcement Learning with Recommendation Sets},
  booktitle = {Proceedings of the 28th International Joint Conference on Artificial Intelligence},
  pages = {2592--2599},
  year = {2019}
}

@inproceedings{zhang2022batchrl,
  author = {Zhang, Qihua and Liu, Junning and Dai, Yuzhuo and Qi, Yiyan and Yuan, Yifan and Zheng, Kunlun and Huang, Fan and Tan, Xianfeng},
  title = {Multi-Task Fusion via Reinforcement Learning for Long-Term User Satisfaction in Recommender Systems},
  booktitle = {Proceedings of the 28th ACM SIGKDD Conference on Knowledge Discovery and Data Mining},
  year = {2022}
}

@inproceedings{zhao2024dt4ier,
  author = {Liu, Ziru and Liu, Shuchang and Zhang, Zijian and Cai, Qingpeng and Zhao, Xiangyu and Zhao, Kesen and Hu, Lantao and Jiang, Peng and Gai, Kun},
  title = {Sequential Recommendation for Optimizing Both Immediate Feedback and Long-Term Retention},
  booktitle = {Proceedings of the 47th International ACM SIGIR Conference on Research and Development in Information Retrieval},
  year = {2024}
}

@inproceedings{xia2026lerl,
  author = {Xia, Chongjun and Peng, Yanchun and Wang, Xianzhi},
  title = {{LLM}-Enhanced Reinforcement Learning for Long-Term User Satisfaction in Interactive Recommendation},
  booktitle = {Database Systems for Advanced Applications},
  series = {Lecture Notes in Computer Science},
  volume = {16535},
  publisher = {Springer},
  year = {2026}
}

@inproceedings{schnabel2016recommendations,
  author = {Schnabel, Tobias and Swaminathan, Adith and Singh, Ashudeep and Chandak, Navin and Joachims, Thorsten},
  title = {Recommendations as Treatments: Debiasing Learning and Evaluation},
  booktitle = {Proceedings of the 33rd International Conference on Machine Learning},
  pages = {1670--1679},
  year = {2016}
}

@inproceedings{wang2019doubly,
  author = {Wang, Xiaojie and Zhang, Rui and Sun, Yu and Qi, Jianzhong},
  title = {Doubly Robust Joint Learning for Recommendation on Data Missing Not at Random},
  booktitle = {Proceedings of the 36th International Conference on Machine Learning},
  pages = {6638--6647},
  year = {2019}
}

@article{gao2023causal,
  author = {Gao, Chen and Zheng, Yu and Wang, Wenjie and Feng, Fuli and He, Xiangnan and Li, Yong},
  title = {Causal Inference in Recommender Systems: A Survey and Future Directions},
  journal = {ACM Transactions on Information Systems},
  volume = {42},
  number = {4},
  year = {2024}
}

@inproceedings{kumar2020cql,
  author = {Kumar, Aviral and Zhou, Aurick and Tucker, George and Levine, Sergey},
  title = {Conservative {Q}-Learning for Offline Reinforcement Learning},
  booktitle = {Advances in Neural Information Processing Systems},
  volume = {33},
  pages = {1179--1191},
  year = {2020}
}

@inproceedings{wu2020mind,
  author = {Wu, Fangzhao and Qiao, Ying and Chen, Jiun-Hung and Wu, Chuhan and Qi, Tao and Lian, Jianxun and Liu, Danyang and Xie, Xing and Gao, Jianfeng and Wu, Winnie and Zhou, Ming},
  title = {{MIND}: A Large-Scale Dataset for News Recommendation},
  booktitle = {Proceedings of the 58th Annual Meeting of the Association for Computational Linguistics},
  pages = {3597--3606},
  year = {2020}
}

@inproceedings{gao2022kuairec,
  author = {Gao, Chongming and Li, Shijun and Lei, Wenqiang and Chen, Jiawei and Li, Biao and Jiang, Peng and He, Xiangnan and Mao, Jiaxin and Chua, Tat-Seng},
  title = {{KuaiRec}: A Fully-Observed Dataset and Insights for Evaluating Recommender Systems},
  booktitle = {Proceedings of the 31st ACM International Conference on Information and Knowledge Management},
  year = {2022}
}

@inproceedings{xie2022cl4srec,
  author = {Xie, Xu and Sun, Fei and Liu, Zhaoyang and Wu, Shiwen and Gao, Jinyang and Ding, Bolin and Cui, Bin},
  title = {Contrastive Learning for Sequential Recommendation},
  booktitle = {2022 IEEE 38th International Conference on Data Engineering},
  pages = {1259--1273},
  year = {2022}
}

@inproceedings{hou2022unisrec,
  author = {Hou, Yupeng and Mu, Shanlei and Zhao, Wayne Xin and Li, Yaliang and Ding, Bolin and Wen, Ji-Rong},
  title = {Towards Universal Sequence Representation Learning for Recommender Systems},
  booktitle = {Proceedings of the 28th ACM SIGKDD Conference on Knowledge Discovery and Data Mining},
  pages = {585--593},
  year = {2022}
}

@inproceedings{zhao2023dt4rec,
  author = {Zhao, Kesen and Zou, Lixin and Zhao, Xiangyu and Wang, Maolin and Yin, Dawei},
  title = {User Retention-Oriented Recommendation with Decision Transformer},
  booktitle = {Proceedings of the ACM Web Conference 2023},
  year = {2023}
}

@inproceedings{he2025llm2rec,
  author = {He, Yingzhi and Liu, Xiaohao and Zhang, An and Ma, Yunshan and Chua, Tat-Seng},
  title = {{LLM2Rec}: Large Language Models Are Powerful Embedding Models for Sequential Recommendation},
  booktitle = {Proceedings of the 31st ACM SIGKDD Conference on Knowledge Discovery and Data Mining},
  year = {2025}
}

@inproceedings{li2026unim,
  title={UniM: A Unified Any-to-Any Interleaved Multimodal Benchmark},
  author={Li, Yanlin and Guo, Minghui and Zhang, Kaiwen and Zhang, Shize and Zhao, Yiran and Li, Haodong and Zhou, Congyue and Zheng, Weijie and Yan, Yushen and Wu, Shengqiong and others},
  booktitle={Proceedings of the IEEE/CVF Conference on Computer Vision and Pattern Recognition},
  pages={15902--15911},
  year={2026}
}
\end{document}